\documentclass[default,iicol,sn-mathphys,Numbered]{sn-jnl}% Default with double column layout

\usepackage{graphicx}%
\usepackage{multirow}%
\usepackage{amsmath,amssymb,amsfonts}%
\usepackage{amsthm}%
\usepackage{mathrsfs}%
\usepackage[title]{appendix}%
\usepackage{xcolor}%
\usepackage{textcomp}%
\usepackage{manyfoot}%
\usepackage{booktabs}%
\usepackage{algorithm}%
\usepackage{algorithmicx}%
\usepackage{algpseudocode}%
\usepackage{listings}%
\usepackage[caption=false, font=footnotesize]{subfig}%
\usepackage{makecell}%
\newcommand{\norm}[1]{\mathinner{\Vert#1\Vert}}%
\newtheorem{definition}{Definition}%

\begin{document}

\title[Energy-Aware Hierarchical Control of Joint Velocities]{Energy-Aware Hierarchical Control of Joint Velocities}


% The affiliation(s) of the author(s), i.e. institution, (department), city, (state), country. A clear indication and an active e-mail address of the corresponding author. If available, the 16-digit ORCID of the author(s).
\author*[1]{\fnm{Jonas} \sur{Wittmann}}\email{jonas.wittmann@tum.de}

\author[1]{\fnm{Daniel} \sur{Hornung}}\email{daniel.hornung@tum.de}

\author[1]{\fnm{Korbinian} \sur{Griesbauer}}\email{korbinian.griesbauer@tum.de}

\author[1]{\fnm{Daniel} \sur{Rixen}}\email{rixen@tum.de}

\affil[1]{Technical University of Munich, TUM School of Engineering and Design, Department of Mechanical Engineering, Chair of Applied Mechanics; Munich Institute of Robotics and Machine Intelligence (MIRMI); \orgaddress{\street{Boltzmannstr. 15}, \city{Garching}, \postcode{85748}, \country{Germany}}}

\abstract{Nowadays, robots are applied in dynamic environments. For a robust operation, the motion planning module must consider other tasks besides reaching a specified pose: (self) collision avoidance, joint limit avoidance, keeping an advantageous configuration, etc. Each task demands different joint control commands, which may counteract each other. We present a hierarchical control that, depending on the robot and environment state, determines online a suitable priority among those tasks. Thereby, the control command of a lower-prioritized task never hinders the control command of a higher-prioritized task. We ensure smooth control signals also during priority rearrangement. Our hierarchical control computes reference joint velocities. However, the underlying concepts of hierarchical control differ when using joint accelerations or joint torques as control signals instead. So, as a further contribution, we provide a comprehensive discussion on how joint velocity control, joint acceleration control, and joint torque control differ in hierarchical task control. We validate our formulation in an experiment on hardware.}%


\maketitle

%
%
% ====================
% --- Introduction ---
% ====================
%
\section{Introduction}\label{sec:Introduction}%
Robots in dynamic environments must consider several tasks simultaneously, e.g. reaching a goal and avoiding obstacles. We define a task as follows:\newline%
\begin{definition}\label{def:task}%
A task is a user-defined behavior of the robot for which a control target $\mathbf{u}(\mathbf{q},\dot{\mathbf{q}})$ needs to be computed in a task space of dimension~$m$. This control depends on the current robot state, i.e. the current joint configuration~$\mathbf{q}$ and joint velocity~$\dot{\mathbf{q}}$, which are expressed in the robot's configuration space ($\mathcal{C}$-space) of dimension~$n$. The task Jacobian \hbox{$\mathbf{J}(\mathbf{q})\in\mathbb{R}^{m\times n}$} maps the control command from $\mathcal{C}$-space to task space. A scalar metric \hbox{$\eta(\mathbf{q})\in[0,~1]$} expresses the current importance of the task. Mathematically formulated:%
\begin{equation}%
    \textsc{task}: \mathbf{q}, \mathbf{\dot{q}} \mapsto \mathbf{J},\mathbf{u},\eta\nonumber%
\end{equation}\end{definition}%
\noindent For readability, we omit the dependency of $\mathbf{J}$, $\mathbf{u}$, and $\eta$ on the robot state in the following. An example for a task is \textit{obstacle avoidance}, which may be formulated as a force that pushes the closest point on the robot away from the obstacle. So $\mathbf{u}\in\mathbb{R}^3$ would be a 3D force vector whose direction is along the closest distance and whose norm is determined by a user-defined law that relates the force~$\mathbf{u}$ to~$\mathbf{q}$ and optionally~$\dot{\mathbf{q}}$, e.g. with a PD control. The task Jacobian $\mathbf{J}(\mathbf{q})\in\mathbb{R}^{3\times n}$ would be the translational Jacobian of the closest point on the robot, and $\eta$ would increase to one the closer the robot approaches the obstacle.\newline%
In the given example, $\mathbf{u}$ is a desired force~$\mathbf{F}_d$ in task space, but the task could also be formulated to compute a desired task space velocity~$\dot{\mathbf{x}}_d$ or acceleration~$\ddot{\mathbf{x}}_d$. In the following, subscript~$\star_d$ represents desired values while values without subscript indicate measured values. So according to~\cite{Siciliano2008}, we distinguish:%
\begin{enumerate}%
    \item In \textit{first-order differential kinematics control}, denoted as \textit{velocity control} in the following, $\mathbf{u}$ is a desired task space \textit{velocity}~$\dot{\mathbf{x}}_d$ that maps to a desired $\mathcal{C}$-space velocity~$\dot{\mathbf{q}}_d$ according to:%
    \begin{equation}\label{Eq:FirstOrder}%
        \dot{\mathbf{q}}_d = \mathbf{J}^{\#}_W\dot{\mathbf{x}}_d%
    \end{equation}%
    $\mathbf{J}^{\#}_W$ is the $\mathbf{W}$-weighted pseudo-inverse of the task Jacobian, and $\mathbf{W}$~is symmetric positive definite:
    \begin{equation}%
        \mathbf{J}^{\#}_W = \mathbf{W}^{-1}\mathbf{J}^T\left(\mathbf{J}\mathbf{W}^{-1}\mathbf{J}^T\right)^{-1}\nonumber%
    \end{equation}%
    \item In \textit{second-order differential kinematics control}, denoted as \textit{acceleration control} in the following, $\mathbf{u}$ is a desired task space \textit{acceleration}~$\ddot{\mathbf{x}}_d$ that maps to a desired $\mathcal{C}$-space acceleration~$\ddot{\mathbf{q}}_d$ according to:%
    \begin{equation}\label{Eq:SecondOrder}%
        \ddot{\mathbf{q}}_d = \mathbf{J}^{\#}_W\left(\ddot{\mathbf{x}}_d-\dot{\mathbf{J}}\dot{\mathbf{q}}\right)%
    \end{equation}%
    \item In \textit{torque control}, $\mathbf{u}$ is a desired task space \textit{force}~$\mathbf{F}_d$ that maps to a desired $\mathcal{C}$-space torque~$\boldsymbol{\tau}_d$ according to:%
    \begin{equation}\label{Eq:DynamicControl}%
        \boldsymbol{\tau}_d = \mathbf{J}^T\mathbf{F}_d%
    \end{equation}%
\end{enumerate}%
Concerning the terminology, we point out that, in torque control, $\mathbf{F}_d$ could be an impedance controller that tracks a desired motion and so, in the end, is used to track a desired task space position, velocity, and acceleration. So, one can argue that in this case, $\boldsymbol{\tau}_d$ is used for velocity control, but we refer to it as torque control. The same holds for velocity control, in which $\dot{\mathbf{x}}_d$ might be the output of an admittance controller and so, in the end, is used to track a desired task space force. So, one can argue that in this case, $\dot{\mathbf{q}}_d$ is used for force control, but we refer to it as velocity control.\newline%
Equations~(\ref{Eq:FirstOrder}-\ref{Eq:DynamicControl}) map a task control to $\mathcal{C}$-space, and we refer to this idea as \textit{task mapping} in the following. These three mappings are linear. This is not the case on position-level, where the mapping between a task space variable $\mathbf{x}_d(\mathbf{q})$ and a corresponding $\mathcal{C}$-space configuration $\mathbf{q}$ may be nonlinear, e.g. in motion tracking, where $\mathbf{x}_d(\mathbf{q})$ is the desired Cartesian pose of the end effector. In offline modules, such as position-level inverse kinematics for path planning, the nonlinear mapping is solved iteratively, e.g. with a Newton-based approach that exploits the linear mapping on velocity-level. As we target online control, we will not consider the mapping on position-level in the following.\newline%
Hierarchical control deals with several tasks that must be accomplished simultaneously, defining priorities among those. It ensures that $\mathcal{C}$-space control commands due to lower-prioritized tasks do not counteract $\mathcal{C}$-space control commands originating from higher-prioritized tasks. It is possible to execute several tasks simultaneously because there are $n$ $\mathcal{C}$-space control variables, but one task may have a task space of lower dimension $m<n$, and so there are $n-m$ remaining degrees of freedom (DoF) to achieve lower-prioritized tasks. That is, the robot is redundant w.r.t. the higher-prioritized task. Hierarchical control uses projection techniques to avoid task conflicts: One can project the lower-prioritized control into the nullspace of a higher-prioritized control. That way, the control of the lower-prioritized task is not fully executed by the robot, but only to such an extend that it has no effect in the higher-prioritized task space. Therefore, the \textit{nullspace projection matrix}~$\mathbf{N}$ is used that computes the nullspace of the higher-prioritized task Jacobian, e.g. for velocity control: \hbox{$\mathbf{N} = \mathbf{I}-\mathbf{J}^{\#}_{W}\mathbf{J}$}. In the following, we refer to the idea of projecting a lower-prioritized task into the nullspace of a higher-prioritized task as \textit{nullspace projection}. For torque control, \cite{Dietrich2015}~discusses the \textit{dynamical consistency} of a nullspace projection as a first important property, and we add the discussion for velocity and acceleration control in Section~\ref{sec:Projection}. Further, \cite{Dietrich2015}~discusses the \textit{strictness} of a nullspace projection as a second important property that is relevant when there are more than two active tasks. They distinguish \textit{successive} and \textit{augmented} projections. The former is non-strict/soft, i.e. the third-level task, though not counteracting the second-level task, may still impact the first-level task. The latter implements a strict hierarchy with no task conflicts on any level. The difference is that in the augmented approach, the task Jacobian of a lower-prioritized task includes all the task Jacobians of higher-prioritized tasks. We use the augmented approach in the following.\newline%
Some hierarchical control approaches define a static hierarchy in which safety constraints are put at the first priority levels and activated only when necessary. Other approaches define a dynamic hierarchy that evolves during robot operation: The example for a \textit{collision avoidance} task given above increases the task priority the closer the robot to a collision. For these approaches with dynamic hierarchy, there are solutions to avoid lower-prioritized tasks counteract higher-prioritized tasks and to ensure that the joint control commands are smooth during task priority rearrangment~\cite{Liu2015, Dehio2019, Dietrich2015}. Our contributions to these existing solutions are:%
\begin{itemize}%
    \item The proposed solutions focus on only one of the three approaches: velocity control, acceleration control, or torque control. But, a comprehensive discussion on how they differ in terms of task mapping characteristics and nullspace projection characteristics was, so far, not reported. We provide this discussion in Section~\ref{sec:Theory}.%
    \item One further contribution consists in an enhanced version of the hierarchical control approach in \cite{Liu2015, Dehio2019}: 1) Our approach allows to consider the energy consumption of the robot; 2) Our approach rearranges task priorities online based on \textit{importance metrics} that consider a robust robot operation. The existing approach defines hierarchy switching points offline based on time, which is not useful for online applications. For example, it is predefined that after 3.2~s, the least prioritized task will become the most prioritized task.%
    \item \cite{Liu2015, Dehio2019} apply the approach in simulations. We apply the approach to hardware demonstrating its real-time capability.%
\end{itemize}%
%
%
% ====================
% --- Related Work ---
% ====================
%
\section{Related Work}%
\cite{Siciliano1991} proposes a framework to handle multiple hierarchies for velocity and acceleration control. However, it enforces a strict hierarchy that leads to discontinuous control signals at priority rearrangements, and it is only verified in simulation. \cite{Mansard2007}~presents a velocity control that prioritizes tasks online. However, they remove tasks based on expected task conflicts, whereas we consider task importance. They validate their approach in experiments on a \textit{Gantry} robot with six DoF and achieve a control cycle of 25~Hz, whereas our robots demand a 1~kHz real-time control. \cite{Salini2011}~introduces non-strict hierarchies and, similar to our approach, uses an automated task reprioritization according to relative importance. However, similar to \cite{Liu2015,Dehio2019}, the change of the relative importance is based on time. Further, they focus on torque control and only validate in simulation. \cite{Dehio2018}~provides further details on~\cite{Dehio2019} and formulates the different nullspace projectors for torque control and velocity control. However, analysis and results are only provided for torque control. The hierarchical control in~\cite{Dietrich2012} ensures continuous control signals during the activation and deactivation of tasks. Therefore, they shape the nullspace projector based on a task activation variable, which is also our approach. They provide experiments on a humanoid robot. However, their presentation is limited to torque control. \cite{Kim2019} applies hierarchical quadratic programming to implement continuous task transitions and insertions. They compute continuous control commands when tasks are inserted, removed, or rearranged, and, as they rely on nonlinear optimization, they can handle differently formulated tasks, i.e. equality and inequality tasks. They provide experiments on a seven DoF robot using kinematic control. However, their approach formulates QP problems online, so they depend on additional third-party software to solve those. They achieve an average computation time of 3.7~ms for a three-level hierarchy. Based on \cite{Liu2015,Dehio2019}, we provide a closed-form solution and achieve an average computation time of less than 1.0~ms for a four-level hierarchy. \cite{Tarbouriech2018}~also presents a QP-based hierarchical velocity control and evaluates the approach with two \textit{KUKA LWR4}. The hierarchy contains a primary motion task and a secondary joint limit avoidance task. \cite{Hu2015}~presents a velocity-based control approach for a dual-arm manufacturing application with two \textit{KUKA LBR iiwa}. A hierarchy of three tasks, including trajectory tracking, joint limit avoidance, and respecting joint velocity limits, is considered. However, the framework allows the adaption of the hierarchy only by removing the lowest task from the Jacobian, which is triggered by rank deficiency. That is, they do not provide a solution to adapt priorities based on task importance. Further, they suffer from discontinuities in the computed velocity control signals. In their conclusion, they outline the importance of investigating an inertia-weighted pseudo-inverse, which we present in our work. Our work builds upon~\cite{Liu2015} and~\cite{Dehio2019}, and we will outline those methods and our modifications throughout the paper.%
%
%
% ====================
% -- Proposed Method -
% ====================
%
\section{Task Mapping, Nullspace Projection, and the Effect of the Weighting Matrix}\label{sec:Theory}%
Task mapping and nullspace projection with their characteristics are well discussed in literature. However, they are often not clearly separated. In this section, we discuss how the selection of the weighting matrix $\mathbf{W}$ in $\mathbf{J}^{\#}_W$ affects the characteristics of task mapping and nullspace projection, and thereby, we distinguish velocity, acceleration, and torque control. We have not found this comprehensive discussion in literature.%
\begin{figure*}%
	\centering%
	\fontsize{8pt}{11pt}\selectfont%
	\subfloat[Four DoF Manipulator in 2D\label{fig:4Dof}]{%
		\includegraphics{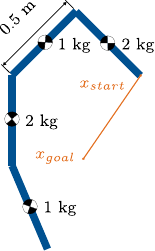}%
	}\hfil%
	\subfloat[Effect of the choice of $\mathbf{W}$ on the kinetic energy $\frac{1}{2}\dot{\mathbf{q}}^T\mathbf{M}\dot{\mathbf{q}}$\label{fig:WeightingMatrix}]{%
		\includegraphics{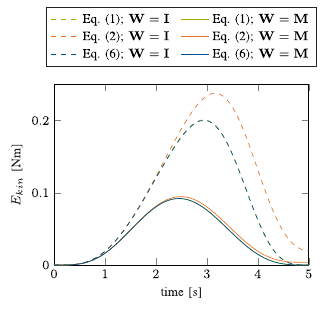}%
	}\hfil%
	\subfloat[Effect of the choice of $\mathbf{W}$ on the dynamical consistency\label{fig:DynamicConistency}]{%
		\includegraphics{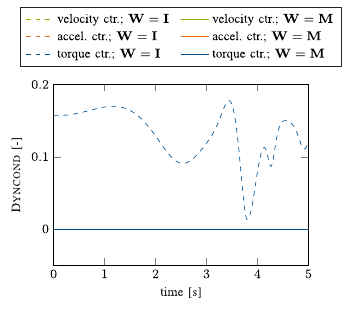}%
	}%
	\caption{A manipulator with four DoF tracks a 2D translational trajectory ($\mathbf{x}_d(t)$,$\dot{\mathbf{x}}_d(t)$,$\ddot{\mathbf{x}}_d(t)$). Figure~\ref{fig:4Dof} shows that in the dynamic robot model, the links are approximated as point masses at the center of mass of each link. For this scenario, Figure~\ref{fig:WeightingMatrix} benchmarks Eqs.~(\ref{Eq:FirstOrder}), (\ref{Eq:SecondOrder}) and (\ref{Eq:OurResolution}) for the case that the only given task is to track the trajectory. The curves corresponding to Eqs.~(\ref{Eq:FirstOrder}) and~(\ref{Eq:OurResolution}), i.e. the green and blue lines, are superposed. Figure~\ref{fig:DynamicConistency} shows the dynamical consistency condition $\textsc{DynCon}$ in Eqs.~(\ref{eq:DynCondDyn}) and~(\ref{eq:DynConKin}) for the case that trajectory tracking is the primary task and the identity joint command~$\mathbf{I}^{n\times 1}$ is used as a secondary task. The curves for $\mathbf{W}=\mathbf{M}$ and for velocity and acceleration control are superposed, i.e. all lines, except the dashed blue line, are superposed}%
	\label{fig:KinAndDynControl}%
\end{figure*}%
\subsection{Optimized Task Mapping for Velocity and Acceleration Control}\label{sec:Mapping}%
First, we investigate task mapping, i.e. we map the task's control from its task space to $\mathcal{C}$-space according to Eqs.~(\ref{Eq:FirstOrder}),~(\ref{Eq:SecondOrder}), or~(\ref{Eq:DynamicControl}). So in this section, we consider only one task, and thus, there is no nullspace projection of any other tasks.\newline%
Torque control of redundant robots defines a unique mapping from task space to $\mathcal{C}$-space with Eq.~(\ref{Eq:DynamicControl}). This is not the case in velocity and acceleration control that have infinite solutions that differ in the selection of $\mathbf{W}$ in \hbox{$\mathbf{J}^{\#}_W$}. Thus, the optimized mapping from task space to $\mathcal{C}$-space, i.e. the optimal choice of $\mathbf{W}$, is only relevant for velocity and acceleration control.\newline%
\hbox{$\mathbf{W}=\mathbf{I}^{n\times n}$} is a typical choice to minimize the local $\text{L}^2$-norm of $\dot{\mathbf{q}}$ in velocity control or $\ddot{\mathbf{q}}$ in acceleration control. This weighting is relevant to keep the $\mathcal{C}$-space control feasible, i.e. within the joint velocity and acceleration limits. For velocity control, it was proposed as \textit{Resolved Motion Rate Control} in~\cite{Whitney1969}.\newline%
Using the $\mathcal{C}$-space mass matrix instead, i.e. \hbox{$\mathbf{W}=\mathbf{M}(\mathbf{q})$}, is another typical choice, and it is usually stated that it minimizes the instantaneous kinetic energy that the robot requires to execute the task control. We drop the dependency of $\mathbf{M}$ on $\mathbf{q}$ in the following for a shorter notation. This is discussed in literature \cite{Dietrich2015,Siciliano2008,Khatib1987}, however, the mappings in Eqs.~(\ref{Eq:FirstOrder}-\ref{Eq:DynamicControl}) are not clearly separated, which is crucial. For velocity control, indeed, \hbox{$\mathbf{W}=\mathbf{M}$} computes the joint velocities for a given task that minimize the instantaneous kinetic energy. This directly results from the first-order optimality conditions of the following optimization problem:%
\begin{alignat}{2}%
    &\!\min_{\dot{\mathbf{q}}}\quad&&\frac{1}{2}\dot{\mathbf{q}}^T\mathbf{W}\dot{\mathbf{q}}\label{eq:ObjFct}\\%
    &\text{s. t.:}&&\dot{\mathbf{x}}_d = \mathbf{J}\dot{\mathbf{q}}\label{eq:Constr}%
\end{alignat}%
However, we want to emphasize that this does not hold for acceleration control: Using \hbox{$\mathbf{W}=\mathbf{M}$} in Eq.~(\ref{Eq:SecondOrder}) tracks the desired task space command but is not the solution that minimizes the kinetic energy. Instead, the acceleration control for minimum instantaneous kinetic energy is derived from Eq.~(\ref{Eq:FirstOrder}):%
\begin{equation}\label{Eq:OurResolution}%
    \ddot{\mathbf{q}}_d = \mathbf{J}^{\#}_W\ddot{\mathbf{x}}_d+\dot{\left(\mathbf{J}^{\#}_W\right)}\mathbf{J}\dot{\mathbf{q}}%
\end{equation}%
For $\dot{\left(\mathbf{J}^{\#}_W\right)}$, often numerically approximated gradients are used, e.g. in~\cite{Nemec1997}. We use the analytical derivative provided in~\cite{Guo2014} and we use the chain rule for the required terms \hbox{$\dot{\mathbf{J}} = \frac{\text{d}\mathbf{J}}{\text{d}\mathbf{q}}\dot{\mathbf{q}}$} and \hbox{$\dot{\mathbf{W}} = \frac{\text{d}\mathbf{W}}{\text{d}\mathbf{q}}\dot{\mathbf{q}}$}. We use the algorithm in~\cite{Haviland2020} to compute~$\frac{\text{d}\mathbf{J}}{\text{d}q_i}$ and we use numeric differentiation to compute~$\frac{\text{d}\mathbf{W}}{\text{d}q_i}$ for~\hbox{$\mathbf{W}=\mathbf{M}$}.\newline%
Figure~\ref{fig:WeightingMatrix} shows the effect of the choice of~$\mathbf{W}$ during task mapping on the kinetic energy consumption. We implemented a manipulator with four links in \textit{MATLAB} (see Figure\ \ref{fig:4Dof}). Its single task is to track a given task space trajectory, which is a two-dimensional point-to-point (PTP) motion with quintic spline interpolation that enforces zero task space velocities and accelerations at start and goal. We compare the three approaches in Eqs.~(\ref{Eq:FirstOrder}),~(\ref{Eq:SecondOrder}) and~(\ref{Eq:OurResolution}).\newline%
The following is obvious: $\mathbf{W}=\mathbf{M}$ minimizes the instantaneous kinetic energy in velocity control. To achieve the same in acceleration control, still $\mathbf{W}=\mathbf{M}$ is required, but Eq.~(\ref{Eq:OurResolution}) must be used instead of Eq.~(\ref{Eq:SecondOrder}).%
\subsection{Dynamically Consistent Nullspace Projection for Torque Control}\label{sec:Projection}%
Section~\ref{sec:Mapping} considered the execution of one task. This section considers the simultaneous execution of several tasks denoted by subscript~$k$, and it shows the impact that $\mathbf{W}$ has on the so-called \textit{dynamical consistency} property of a nullspace projection. As outlined in Section~\ref{sec:Introduction}, the $\mathcal{C}$-space commands due to the lower-prioritized task~$k+1$ are projected into the nullspace of the higher-prioritized task~$k$ with the nullspace projector~$\mathbf{N}_{k+1}$. The projector for torque control $\mathbf{N}_{tor, k+1}$ is the transposed of the projector for velocity and acceleration control~$\mathbf{N}_{vel, k+1}=\mathbf{N}_{acc, k+1}$:%
\begin{align}
    \mathbf{N}_{vel, k+1} &= \mathbf{N}_{acc, k+1} &&= \mathbf{I} - \mathbf{J}^{\#}_{W,k}\mathbf{J}_k\label{Eq:KinProj}\\%
    \mathbf{N}_{tor, k+1} &= \mathbf{N}_{vel, k+1}^T &&= \mathbf{I} - \mathbf{J}^{T}_k\left(\mathbf{J}^{\#}_{W,k}\right)^T\label{Eq:DynProj}%
\end{align}%
When choosing $\mathbf{W}=\mathbf{I}$, $\mathbf{N}_{vel}$ is symmetric and thus \hbox{$\mathbf{N}_{vel} = \mathbf{N}_{tor}$} holds. However, this is not the case for \hbox{$\mathbf{W}=\mathbf{M}$}.\newline%
\cite{Dietrich2015}~discusses the computation of dynamically consistent nullspace projections for torque control. Dynamically consistent means that a lower-prioritized task does not generate accelerations in a higher-prioritized task space. \cite{Dietrich2015}~formulates this condition, which we call $\textsc{DynCon}$, for torque control, and we add the condition for velocity and acceleration control:%
\begin{alignat}{3}%
    &\textsc{DynCon}_{tor}: &&\norm{\mathbf{J}_k\mathbf{M}^{-1}\mathbf{N}_{k+1}\boldsymbol{\tau}_{k+1}}_2 && = 0\label{eq:DynCondDyn}\\%
    &\textsc{DynCon}_{vel/acc}: &&\norm{\mathbf{J}_k\mathbf{N}_{k+1}\mathbf{\ddot{q}}_{k+1}}_2 && = 0\label{eq:DynConKin}%
\end{alignat}%
For torque control, \cite{Dietrich2015} proves that selecting \hbox{$\mathbf{W}=\mathbf{M}$} for the nullspace projector in Eq.~(\ref{Eq:DynProj}) ensures dynamical consistency while $\mathbf{W}=\mathbf{I}$ does not. Velocity and acceleration control inherently implement dynamically consistent projections, i.e. a lower-prioritized task never generates accelerations in a higher-prioritized task independent from $\mathbf{W}$. The dynamical consistency property is, therefore, only relevant for torque control. Note that the term dynamical consistency is also used in velocity and acceleration control when \hbox{$\mathbf{W}=\mathbf{M}$}~\cite{Dietrich2013}. However, in these applications \hbox{$\mathbf{W}=\mathbf{I}$} also fulfills Eq.~(\ref{eq:DynConKin}).\newline%
Figure~\ref{fig:DynamicConistency} shows the left side of $\textsc{DynCon}$ in Eqs.~(\ref{eq:DynCondDyn}) and~(\ref{eq:DynConKin}). The scenario is the same as in Section~\ref{sec:Mapping}. However, now, there is a certain secondary task. Here, we simply assume that this secondary task requests unit commands, i.e. $\dot{\mathbf{q}}_d=\mathbf{I}^{n\times 1}~\frac{\text{rad}}{\text{s}}$, $\ddot{\mathbf{q}}_d=\mathbf{I}^{n\times 1}~\frac{\text{rad}}{\text{s}^2}$, or $\boldsymbol{\tau}_d=\mathbf{I}^{n\times 1}~\text{Nm}$. For example, in velocity control, the secondary task sets a target velocity of 1~rad/s in each joint, or, in dynamic control, the secondary task sets a target torque of 1~Nm in each joint. We want to investigate the dynamical consistency, i.e. if these secondary task controls generate accelerations in the task space of the primary task.\newline%
The following is obvious in Figure~\ref{fig:DynamicConistency}: $\mathbf{W}=\mathbf{M}$ ensures a dynamically consistent projection in torque control, and $\mathbf{W}=\mathbf{I}$ does not. Projections in velocity and acceleration control are dynamically consistent independent of~$\mathbf{W}$.\newline%
Note that in Section~\ref{sec:Mapping}, we defined the 2D translational trajectory ($\mathbf{x}_d(t)$,$\dot{\mathbf{x}}_d(t)$,$\ddot{\mathbf{x}}_d(t)$). So, using (\ref{Eq:FirstOrder}),~(\ref{Eq:SecondOrder}), and~(\ref{Eq:OurResolution}), we could compute the desired joint velocities and the desired joint acceleration to track the trajectory. To track this trajectory with dynamic control, i.e. with joint torques $\boldsymbol{\tau}_d=\mathbf{J}^T\mathbf{F}_d$, we formulate the dynamics in task space according to~\cite{Siciliano2008b}:%
\begin{alignat}{2}%
    & \overline{\mathbf{M}} &&= \left(\mathbf{J}\mathbf{M}^{-1}\mathbf{J}^T\right)^{-1}\\%
    & \overline{\mathbf{c}} &&= \overline{\mathbf{M}}\mathbf{J}\mathbf{M}^{-1}\mathbf{C}\dot{\mathbf{q}} - \overline{\mathbf{M}}\dot{\mathbf{J}}\dot{\mathbf{q}}\\%
    & \overline{\mathbf{g}} &&= \overline{\mathbf{M}}\mathbf{J}\mathbf{M}^{-1}\mathbf{g}\\%
    &\mathbf{F}_d &&= \overline{\mathbf{M}}\ddot{\mathbf{x}}_d+\overline{\mathbf{c}}+\overline{\mathbf{g}}%
\end{alignat}%
$\mathbf{M}$, $\mathbf{C}$ and $\mathbf{g}$ are the mass matrix, Coriolis and centrifugal terms, and gravity torques respectively formulated in $\mathcal{C}$-space. $\overline{\mathbf{M}}$, $\overline{\mathbf{c}}$ and $\overline{\mathbf{g}}$ are the corresponding terms formulated in task space.%
\subsection{Task Mapping and Nullspace Projection in Velocity Control}\label{sec:MappingAndProjection}%
\begin{table*}[tb]%
	\centering%
	\caption{Capabilities of Eqs.~(\ref{Eq:ASC}) and~(\ref{Eq:RealSolution}) to minimize kinetic energy consumption and error of secondary task}%
	\begin{tabular}[t]{rlccc}%
		\hline%
		Line & Approach & $\overline{E}_{kin}$ & $\overline{L}$ & $\overline{E}_{kin}+\overline{L}$\\%
		\hline%
		1: & Eq.~(\ref{Eq:RealSolution}) with $\mathbf{D}=\mathbf{M}$; $\mathbf{E}=\mathbf{I}$ & 104 & 267 & \textbf{371}\\%
	    2: & Eq.~(\ref{Eq:RealSolution}) with $\mathbf{D}=\mathbf{M}$; $\mathbf{E}=\mathbf{M}$ & 99 & 290 & 389\\%
	    3: & Eq.~(\ref{Eq:RealSolution}) with $\mathbf{D}=\mathbf{I}$; $\mathbf{E}=\mathbf{I}$ & 187 & 252 & 439\\%
	    4: & Eq.~(\ref{Eq:RealSolution}) with $\mathbf{D}=\mathbf{M}$; $\mathbf{E}=\mathbf{0}$ & \textbf{84} & 324 & 407\\%
	    5: & Eq.~(\ref{Eq:RealSolution}) with $\mathbf{D}=\mathbf{0}$; $\mathbf{E}=\mathbf{I}$ & 177 & \textbf{236} & 413\\%
	    \hline%
		6: & Eq.~(\ref{Eq:alphaASC}) with $\mathbf{W}=\mathbf{M}$; $\alpha=1.0$ & 116 & 280 & 396\\%
		7: & Eq.~(\ref{Eq:alphaASC}) with $\mathbf{W}=\mathbf{M}$; $\alpha=\frac{2}{3}$ & 99  & 290 & 389\\%
		8: & Eq.~(\ref{Eq:alphaASC}) with $\mathbf{W}=\mathbf{I}$; $\alpha=1.0$ & 177 & \textbf{236} & 413\\%
		9: & Eq.~(\ref{Eq:alphaASC}) with $\mathbf{W}=\mathbf{I}$; $\alpha=\frac{2}{3}$ & 186  & 252 & 439\\%
		10: & Eq.~(\ref{Eq:alphaASC}) with $\mathbf{W}_{\text{map}}=\mathbf{M}$; $\mathbf{W}_{\text{proj}}=\mathbf{I}$; $\alpha=1.0$ & 106  & 268 & 375\\%
		\hline%
	\end{tabular}%
	\label{tbl:ASC}%
\end{table*}%
This section combines task mapping as discussed in Section~\ref{sec:Mapping} and nullspace projection as discussed in Section~\ref{sec:Projection}, and it applies their combination to velocity control. Our contribution consists in proposing a weighting approach that computes a dynamically consistent projection, which trades off energy consumption and tracking error of the lower-prioritized task.\newline%
For velocity control and in the context of task mapping, Section~\ref{sec:Mapping} states that $\mathbf{W}=\mathbf{M}$ ensures minimum instantaneous kinetic energy. For velocity control and its nullspace projection, Section~\ref{sec:Projection} states that projections are dynamically consistent independent of $\mathbf{W}$. So, considering two tasks and combining task mapping and nullspace projection, the final joint velocity control is:%
\begin{align}
    &\dot{\mathbf{q}} = \underbrace{\mathbf{J}^{\#}_{W_{\text{map},1}}\dot{\mathbf{x}}_1}_{\mathbf{\dot{q}}_{1}} + \left(\mathbf{I} - \mathbf{J}^{\#}_{W_{\text{proj},1}}\mathbf{J}_1\right)\underbrace{\mathbf{J}^{\#}_{W_{\text{map},2}}\dot{\mathbf{x}}_2}_{\mathbf{\dot{q}}_{2}}\label{Eq:ASC}\\%
    &\mathbf{W}_{\text{map},1} = \mathbf{W}_{\text{map},2} = \mathbf{M}\\%
    &\mathbf{W}_{\text{proj},1} = \text{arbitrary}%
\end{align}%
That is, the weighting matrices for task mapping $\mathbf{W}_{\text{map}}$ and nullspace projection $\mathbf{W}_{\text{proj}}$ may be chosen independently. Equation~(\ref{Eq:ASC}) is a popular version for velocity control of redundant robots in literature where \hbox{$\mathbf{W}_{\text{map}}=\mathbf{W}_{\text{proj}}=\mathbf{W}$} is commonly used, i.e. task mapping and nullspace projection use the same weighting.\newline%
For only one task, Eq.~(\ref{Eq:FirstOrder}) minimizes the nonlinear optimization in Eqs.~(\ref{eq:ObjFct}-\ref{eq:Constr}), and as outlined, setting $\mathbf{W}=\mathbf{M}$ in Eqs.~(\ref{Eq:FirstOrder},\ref{eq:ObjFct},\ref{eq:Constr}) minimizes instantaneous kinetic energy. Similarly, for two tasks, we formulate the minimization problem:%
\begin{alignat}{2}%
	&\!\min_{\dot{\mathbf{q}}}\quad && \frac{1}{2}\dot{\mathbf{q}}^T\mathbf{D}\dot{\mathbf{q}}+\left(\dot{\mathbf{q}}_2-\dot{\mathbf{q}}\right)^T\mathbf{E}\left(\dot{\mathbf{q}}_2-\dot{\mathbf{q}}\right)\label{Eq:Opt}\\%
	&\text{s. t.:}              &&	\dot{\mathbf{x}}_1-\mathbf{J}_1\dot{\mathbf{q}} = \mathbf{0}%
\end{alignat}%
That is, we look for a compromise between the $\frac{\mathbf{D}}{2}$-weighted squared norm of the joint velocities, corresponding to kinetic energy for $\mathbf{D}=\mathbf{M}$, and the $\mathbf{E}$-weighted tracking error of the secondary task command. The optimality conditions give:%
\begin{align}%
	&\dot{\mathbf{q}} = \mathbf{J}^{\#}_{W_1}\dot{\mathbf{x}}_1 + \left(\mathbf{I} - \mathbf{J}^{\#}_{W_1}\mathbf{J}_1\right)\mathbf{W}_1^{-1}2\mathbf{E}\dot{\mathbf{q}}_2\label{Eq:RealSolution}\\%
	&\mathbf{W}_1 = \mathbf{D} + 2\mathbf{E}\label{Eq:RealCond}%
\end{align}%
In contrast to our solution in Eq.~(\ref{Eq:RealSolution}), one can show that Eq.~(\ref{Eq:ASC}) with \hbox{$\mathbf{W}_{\text{map},1}=\mathbf{W}_{\text{proj},1}=\mathbf{W}_1$} solves:%
\begin{alignat}{2}%
    &\!\min_{\dot{\mathbf{q}}}\quad&&\left(\dot{\mathbf{q}}_2-\dot{\mathbf{q}}\right)^T\mathbf{W}_1\left(\dot{\mathbf{q}}_2-\dot{\mathbf{q}}\right)\label{Eq:Opt_wrong2}\\%
   	&\text{s. t.:}              &&	\dot{\mathbf{x}}_1-\mathbf{J}_1\dot{\mathbf{q}} = \mathbf{0}%
\end{alignat}%
Thus, the commonly used Eq.~(\ref{Eq:ASC}) minimizes the $\mathbf{W}_1$-weighted tracking error only and does not formulate a compromise between kinetic energy and tracking error. However, our solution in Eq.~(\ref{Eq:RealSolution}) allows to formulate this compromise.\newline%
Note that, in Eq.~(\ref{Eq:RealSolution}), if $\mathbf{D}$ is a multiple of $\mathbf{E}$, i.e. $\mathbf{D}=\beta\mathbf{E}$ with $\beta \in \mathbb{R}$, the following holds:%
\begin{equation}%
    \dot{\mathbf{q}} = \mathbf{J}^{\#}_{W_1}\dot{\mathbf{x}}_1 + \frac{2}{\beta+2}\left(\mathbf{I}-\mathbf{J}^{\#}_{W_1}\mathbf{J}_1\right)\dot{\mathbf{q}}_2%
\end{equation}%
This motivates that, in literature, the second term in Eq.~(\ref{Eq:ASC}), i.e. the nullspace projection, is often applied with an additional scalar $\alpha$~\cite{Schuetz2014}:%
\begin{equation}%
    \dot{\mathbf{q}} = \mathbf{J}^{\#}_{W_{\text{map}}}\dot{\mathbf{x}}_1 + \alpha\left(\mathbf{I} - \mathbf{J}^{\#}_{W_{\text{proj}}}\mathbf{J}_1\right)\dot{\mathbf{q}}_2\label{Eq:alphaASC}%
\end{equation}%
For the case that our solution in Eq.~(\ref{Eq:RealSolution}) uses $\mathbf{D}=\beta\mathbf{E}$ and $\alpha=\frac{2}{\beta+2}$ in Eq.~(\ref{Eq:alphaASC}), both approaches are equivalent.\newline%
Table~\ref{tbl:ASC} benchmarks Eqs.~(\ref{Eq:RealSolution}) and~(\ref{Eq:alphaASC}) w.r.t. their capability to minimize kinetic energy and the error of the secondary task command. We use the robot shown in Figure~\ref{fig:4Dof}. The high-priority task is to track a 2D PTP path in task space that we interpolate with a quintic polynomial as in Section~\ref{sec:Mapping}. The low-priority task is to favor a comfort configuration $\mathbf{q}_{cmf}$. % = \left[ \pi/8, -\pi/8, -\pi/4, -\pi/2 \right]^T$.
This task is formulated with gradient descent, i.e. we minimize \hbox{$L = \kappa\frac{1}{2}\norm{\mathbf{q}_{cmf}-\mathbf{q}}_2^2$} with $\kappa=0.1$ by applying the secondary control $\dot{\mathbf{q}}_2 = \mathbf{J}^{\#}_{W_{map}}\left(-\frac{dL}{d\mathbf{q}}\right) = \mathbf{J}^{\#}_{W_{map}}\kappa\left(\mathbf{q}_{cmf}-\mathbf{q}\right)$ with $\mathbf{J}_{2} = \mathbf{I}$. $\kappa$ scales $L$ w.r.t. $E_{kin}$. For Eq.~(\ref{Eq:alphaASC}), except in line~10 of Table~\ref{tbl:ASC}, we use \hbox{$\mathbf{W}_{\text{map}}=\mathbf{W}_{\text{proj}}=\mathbf{W}$}. To get a scenario-independent benchmark, we simulate 100 PTP motions that start at a random joint configuration and reach a randomly sampled goal pose in the reachable workspace. Eqs.~(\ref{Eq:RealSolution}) and~(\ref{Eq:alphaASC}) are applied to the same 100 simulation scenarios. Table~\ref{tbl:ASC} shows the average kinetic energy~$\overline{E}_{kin}$ and the average secondary objective cost~$\overline{L}$ over the 100 simulations. $E_{kin}$ and $L$ of one simulated motion are computed as the sum of uniformly distributed time samples over the trajectory. Note that both approaches minimize the kinetic energy and tracking error only locally. However, for many trajectory simulations, Table~\ref{tbl:ASC} supports our expectation that an approach that is superior w.r.t. the local optimization capability, in the long run, is also superior w.r.t. the global optimization capability.\newline%We evaluate the sum~$\overline{E}_{kin}+\overline{L}$ in Table~\ref{tbl:ASC} because, for $\mathbf{D}=\mathbf{M}$ and $\mathbf{E}=\mathbf{I}$, this is the objective function of the minimization problem in Eq.~(\ref{Eq:Opt}). However, with our approach in Eq.~(\ref{Eq:RealSolution}), it is straightforward, to prioritize for example~$\overline{L}$ over~$\overline{E}_{kin}$ by multiplying~$\mathbf{E}$ with a scaling factor~$\kappa$. For instance, minimizing the square of the velocity norm can be used to avoid violations of the actuator constraints. So the following discussion would also apply to, e.g., $\overline{E}_{kin}+\kappa\overline{L}$.\newline%
From Table~\ref{tbl:ASC}, the following is obvious: Eq.~(\ref{Eq:RealSolution}) with $\mathbf{D}=\mathbf{M}$ and $\mathbf{E}=\mathbf{I}$ minimizes $\overline{E}_{kin}+\overline{L}$ best. As outlined, the approach in Eq.~(\ref{Eq:alphaASC}) with $\alpha=1.0$ only considers the secondary objective, and thus for $\mathbf{W}=\mathbf{I}$ it minimizes~$\overline{L}$ best. This is equivalent to Eq.~(\ref{Eq:RealSolution}) with $\mathbf{D}=\mathbf{0}$ and $\mathbf{E}=\mathbf{I}$ (line 5). As stated, Eq.~(\ref{Eq:RealSolution})~with $\mathbf{D}=\beta\mathbf{E}$, e.g. $\beta=1.0$ in line~2, is equivalent to Eq.~(\ref{Eq:alphaASC}) when $\alpha=\frac{2}{\beta+2}$, e.g. $\alpha=\frac{2}{3}$ in line~7. With our solution in Eq.~(\ref{Eq:RealSolution}), one can scale~$\mathbf{D}$ or~$\mathbf{E}$ to increase or decrease the importance of $E_{kin}$ over $L$, and we show the two extreme cases that scale one of the two to zero (lines~4 and~5). That way, we get the solution with minimum energy and the solution with minimum tracking error. There is the suspicion that by using different weighting matrices in Eq.~(\ref{Eq:alphaASC}) and setting $\mathbf{W}_{\text{map}}=\mathbf{M}$ and $\mathbf{W}_{\text{proj}}=\mathbf{I}$, one gets the solution that trade-offs kinetic energy and tracking error. Table~\ref{tbl:ASC} shows that this gets close to the solution of Eq.~(\ref{Eq:RealSolution}), but still is slightly worse.\newline%
We conclude the following:%
\begin{itemize}
    \item The standard approach in Eq.~(\ref{Eq:ASC}) with \hbox{$\mathbf{W}_{\text{map}}=\mathbf{W}_{\text{proj}}=\mathbf{W}$} minimizes the $\mathbf{W}$-weighted error norm of the secondary task command. Adding $\alpha$ allows to additionally consider the $\mathbf{W}$-weighted velocity norm.%
    \item The standard approach in Eq.~(\ref{Eq:ASC}) with \hbox{$\mathbf{W}_{\text{map}}\neq\mathbf{W}_{\text{proj}}$} implements a strict hierarchy but does not allow a physical interpretation.%
    \item Our approach in Eq.~(\ref{Eq:RealSolution}), in contrast to Eq.~(\ref{Eq:ASC}), allows to consider kinetic energy and secondary task with differently shaped matrices. For instance, using $\mathbf{D}=\mathbf{M}$ and $\mathbf{E}=\mathbf{I}$ finds a compromise between kinetic energy and secondary task tracking.%
\end{itemize}%
%
%
% ====================
% -- Proposed Method -
% ====================
%
\section{Hierarchical Velocity Control}\label{sec:HierarchicalControl}%
Section~\ref{sec:MappingAndProjection}~investigated the combination of task mapping and nullspace projection for velocity control. This section incorporates both into a generic formulation for the hierarchical control of joint velocities. Thereby, the algorithm has to tackle the two challenges outlined in Section~\ref{sec:Introduction}, i.e. ensuring smooth control signals during task rearrangement, and avoiding that lower-prioritized tasks counteract higher-prioritized tasks. Note that the latter, i.e. a strict hierarchy, is only demanded in periods without task rearrangement: During the continuous priority rearrangement, the control computes only soft hierarchies~\cite{Liu2015,Dehio2019}. We aim for a priority rearrangement based on the importance metric $\eta_k$. Based on the investigations in Section~\ref{sec:MappingAndProjection}, we use Eq.~(\ref{Eq:RealSolution}) with $\mathbf{D}=\mathbf{M}$ and $\mathbf{E}=\mathbf{I}$ and thus the projected command of a task $k$ is defined as:%
\begin{equation}%
    \dot{\mathbf{q}}_k = \left(\mathbf{I} - \mathbf{J}^{\#}_{M+2I,k}~\mathbf{J}_k\right)\left(\mathbf{M}+2\mathbf{I}\right)^{-1}2\mathbf{J}^{\#}_{M+2I,k}\mathbf{u}_k\nonumber%
\end{equation}%
Our work builds upon~\cite{Liu2015} and~\cite{Dehio2019}. \cite{Liu2015}~proposes \textit{Generalized Hierarchical Control (GHC)} as a control algorithm for multiple hierarchies. GHC solves the two challenges, i.e. smooth signals and strict hierarchies, only for torque control and further, the derivation of the projector does not consider~$\mathbf{W}$. Thus, GHC can only be used if \hbox{$\mathbf{W}=\mathbf{I}$}. \cite{Dehio2019}~enhances GHC and proposes \textit{Dynamically Consistent GHC (DynGHC)} for torque control. They achieve this by considering~$\mathbf{W}$ in the derivation of the nullspace projector.\newline%
Alg.~\ref{Alg:HierarchicalControl} shows our hierarchical control that is based on~\cite{Liu2015} and~\cite{Dehio2019}. We refer to them for details, but we briefly outline each step and we show our modifications in the following.\newline%
\begin{algorithm}%
	\caption{Hierarchical Velocity Control of $N$ Tasks}%
	\begin{algorithmic}[1]%
		\renewcommand{\algorithmicrequire}{\textbf{Input:}}%
		\renewcommand{\algorithmicensure}{\textbf{Output:}}%
		\Require $\mathbf{q}, \dot{\mathbf{q}},\mathbf{D},\mathbf{E},\mathbf{W}=\mathbf{D}+2\mathbf{E}$%
		\Ensure $\mathbf{\dot{q}}_d$%
		\For {$k \in {1, \dots, N}$}%
		    \State $[\mathbf{J}_k, \mathbf{u}_k, \eta_k] = \text{computeTask}(\mathbf{q},\dot{\mathbf{q}})$%
		\EndFor%
		\State $\mathbf{A} = \text{computeTaskPriorityMatrix}(\eta_1, \dots, \eta_N)$%
		\State $\mathbf{J}_{aug} = \text{computeAugJacobi}(\mathbf{J}_1, \dots, \mathbf{J}_N)$%
		\For {$k \in {1, \dots, N}$}%
		    \State $\mathbf{J}_{aug,k} = \text{getTaskAugJacobi}(\mathbf{A}(k,:),\mathbf{J}_{aug})$%
		    \State $[\mathbf{L},\mathbf{Q},\mathbf{A}_{sr}] = \text{dynGHC}(\mathbf{A}(k,:),\mathbf{J}_{aug,k},\mathbf{W})$%
		    \State $\mathbf{N}_{kin, k} = \mathbf{L}\left(\mathbf{I}-\mathbf{Q}\mathbf{A}_{sr}^T\mathbf{Q}^T\right)\mathbf{L}^{-1}$%
		\EndFor%
		\For {$k \in {1, \dots, N}$}%
		    \State $\mathbf{K}=\eta_k\mathbf{I}+\left(1-\eta_k\right)\left(\mathbf{D}+2\mathbf{E}\right)^{-1}2\mathbf{E}$%
		    \State $\mathbf{\dot{q}}_d \leftarrow \mathbf{\dot{q}}_d + \mathbf{N}_{kin,k}\mathbf{K}\mathbf{J}^{\#}_{W,k}\mathbf{u}_k$%
		\EndFor%
	\end{algorithmic}%
	\label{Alg:HierarchicalControl}%
\end{algorithm}%
\begin{table*}[tb]%
	\centering%
	\caption{Benchmark of GHC, DynGHC and our approach}%
	\begin{tabular}[t]{rlccc}%
		\hline%
		Line & Approach & $\overline{E}_{kin}$ & $\overline{L}$ & $\overline{E}_{kin}+\overline{L}$\\%
		\hline%
		1: & Ours with $\mathbf{D}=\mathbf{M}$; $\mathbf{E}=\mathbf{I}$ & 104 & 267 & \textbf{371}\\%
		2: & Ours with $\mathbf{D}=\mathbf{M}$; $\mathbf{E}=\mathbf{0}$ & \textbf{84} & 324 & 407\\%
		3: & Ours with $\mathbf{D}=\mathbf{0}$; $\mathbf{E}=\mathbf{M}$ & 116 & 280 & 396\\%
		4: & Ours with $\mathbf{D}=\mathbf{0}$; $\mathbf{E}=\mathbf{I}$ & 177 & \textbf{236} & 413\\%
		\hline%
		5: & \textit{DynGHC} with $\mathbf{W}=\mathbf{M}$ & 116 & 280 & 396\\%
		6: & \textit{DynGHC} with $\mathbf{W}=\mathbf{I}$ & 177 & \textbf{236} & 413\\%
		\hline%		
	    7: & \textit{GHC} with $\mathbf{W}=\mathbf{I}$ & 177 & \textbf{236} & 413\\%		
	    \hline%
	\end{tabular}%
	\label{tbl:Proj}%
\end{table*}%
Line~2 computes our defined tasks according to Definition~\ref{def:task}. Line~4 computes the task priority matrix $\mathbf{A}$ that, given the individual importance metrics $\eta_k$, expresses the priorities among all tasks. This builds upon~\cite{Liu2015}, however, they prioritize the tasks based on time, whereas we prioritize based on importance metrics. Line~5, similar to \cite{Dehio2019}, computes the \textit{augmented Jacobian} $\mathbf{J}_{aug}$ according to~\cite{Dietrich2015} that stacks all task Jacobians $\mathbf{J}_k$. In line~6 and the following, we compute the corresponding task projector for each task. To do so, line~7 first rearranges the task Jacobians in $\mathbf{J}_{aug}$ according to the hierarchy in~$\mathbf{A}$. Line~8 computes DynGHC as given in~\cite{Dehio2019}. We refer to~\cite{Dehio2019} for details specifying only here that: $\mathbf{L}$ is the lower triangular matrix of $\mathbf{W} = \mathbf{D}+2\mathbf{E}$ (\cite{Dehio2019} uses $\mathbf{W}=\mathbf{M}$ instead) that stems from a Cholesky decomposition; $\mathbf{Q}$ results from a QR decomposition of $\mathbf{J}_{aug,k}\mathbf{L}$; and $\mathbf{A}_{sr}$ is a diagonal matrix with the priorities of task $k$ among each task which are stored in the $k$-th row of $\mathbf{A}$. Line~9 computes the transposed of the dynamically consistent nullspace projector. We use the transposed version as, in contrast to~\cite{Dehio2019}, we use velocity control. Line~11 maps the task control~$\mathbf{u}_k$ to $\mathcal{C}$-space and projects this command into the nullspace of the higher-prioritized tasks before adding it to the final control $\dot{\mathbf{q}}_d$. However, in line~13 the factor $(\mathbf{D}+2\mathbf{E})^{-1}2\mathbf{E}$ is not directly applied to the control. Instead, line~12 defines a linear interpolation between this factor and identity: Equation~(\ref{Eq:RealSolution}) was derived for a two-level hierarchy in which only the command for the low priority task is scaled with $(\mathbf{D}+2\mathbf{E})^{-1}2\mathbf{E}$. Therefore, we remove that scaling for high-priority tasks for which $\eta_k=1.0$ holds. This also ensures smooth control signals during task reprioritization.\newline%
Table~\ref{tbl:Proj} benchmarks our approach in Alg.~\ref{Alg:HierarchicalControl} with \textit{GHC}~\cite{Liu2015} and \textit{DynGHC}~\cite{Dehio2019}. As we use velocity control, we use the transposed projectors in our formulations of \textit{GHC} and \textit{DynGHC}. We use the same simulation scenario as for Table~\ref{tbl:ASC} with the same 100 random PTP motions. Again, the motion task is of higher priority than keeping the comfort configuration. Later, we will introduce dynamic hierarchies.\newline%
The following is obvious: Our approach in Alg.~\ref{Alg:HierarchicalControl} with $\mathbf{D}=\mathbf{M}$ and $\mathbf{E}=\mathbf{I}$ minimizes $\overline{E}_{kin}+\overline{L}$ best. Using $\mathbf{D}=\mathbf{0}$, it is equivalent to \textit{DynGHC}. As expected, \textit{DynGHC} with $\mathbf{W}=\mathbf{I}$ is equivalent to \textit{GHC}. Note that the results of Table~\ref{tbl:Proj}, computed with Alg.~\ref{Alg:HierarchicalControl}, correspond to the results of Table~\ref{tbl:ASC}, computed with the analytical formula of Section~\ref{sec:MappingAndProjection}: For instance, line~1 of both tables have the same settings and also the results are the same as we applied the same 100 random motions to both. This proves that our implementation in Alg.~\ref{Alg:HierarchicalControl} is correct.%
\subsection{Stack of Tasks}\label{sec:Tasks}%
So far, tracking a precomputed trajectory and keeping a comfort configuration were used for demonstration in 2D. We use the following four tasks in our experiments on real robots, that are implemented with potential fields as introduced in~\cite{Khatib1986} for obstacle avoidance.%
\subsubsection{Motion}%
For hierarchical control, offline computed trajectories, such as the quintic interpolation in Figure~\ref{fig:KinAndDynControl}, are not suitable: Due to task rearrangements, the trajectory can not be tracked at all times with the highest priority, and thus, the robot motion deviates from the offline computed trajectory. Therefore, we use an online trajectory generation approach that is based on potential fields. At any time, the desired end effector velocity~$\dot{\mathbf{x}}_{motion}$ is heading towards the attracting goal pose. So, the direction of~$\dot{\mathbf{x}}_{motion}$ is $\frac{\mathbf{x}_{goal} - \mathbf{x}}{\norm{\mathbf{x}_{goal} - \mathbf{x}}_2}$, and its magnitude is the predefined maximum task space velocity~$\dot{x}_{max}$. In practice, we have to scale down~$\dot{x}_{max}$ at the beginning of the application to avoid a discontinuous velocity jump from the rest position. Therefore, we introduce the scaling~$\gamma(d)$. It scales~$\dot{x}_{max}$ to zero when the robot's distance~$d$ to the start or goal of the trajectory is within the breaking distance $d_{break}$ (else $\gamma = 1.0$). That way, there are no discontinuities in the desired velocity profile $\dot{\mathbf{x}}_{motion}$:%
\begin{align}%
    \dot{\mathbf{x}}_{motion} &= \gamma(d)~\dot{x}_{max}\frac{\mathbf{x}_{goal} - \mathbf{x}}{\norm{\mathbf{x}_{goal} - \mathbf{x}}_2}\\%
    \gamma(d) &= \frac{1}{2}\left(1-\text{cos}\left(\pi\frac{d}{d_{break}}\right)\right)\label{Eq:scaling}%
\end{align}%
At any time, we set $\eta_{motion}=1.0$ because reaching the goal is of high importance independent of the robot state. $\mathbf{J}_{motion}$ is the end effector Jacobian. \textit{Motion} is a six-dimensional task.%
\subsubsection{Joint Limit Avoidance (Jla)} \textit{Jla} pushes the joints away from their limits. The formulation of $\dot{\mathbf{x}}_{Jla}\in\mathbb{R}^n$ is straightforward and given in~\cite{Wittmann2022a}. This task gets activated for joints that are within an activation threshold to their limits. $\mathbf{J}_{Jla}$ is a diagonal matrix with 1.0 for joints within the threshold and 0.0 else. $\eta_{Jla}$ linearly increases from 0.0 to 1.0 within the threshold, i.e. $Jla$ gets more important the closer to the limits.%
\subsubsection{Collision Avoidance (Coll)}%
\textit{Coll} pushes the closest point on the robot to the environment/to the other robot away from the approaching collision. $\dot{\mathbf{x}}_{Coll}\in\mathbb{R}^3$ and $\mathbf{J}_{Coll}$ is the translational Jacobian of the closest point on the robot. \cite{Wittmann2022b}~gives implementation details. Similar to \textit{Jla}, \textit{Coll} increases linearly within an activation distance.%
\subsubsection{Manipulability (Mnp)}%
\textit{Mnp} is based on the manipulability measure in~\cite{Yoshikawa1985} and maximizes $l_{Mnp} = \sqrt{\text{det}(\mathbf{J}\mathbf{J}^{T})}$. We use $\frac{\text{d}l_{Mnp}}{\text{d}\mathbf{q}}$, derived in \cite{Baur2012}, as the desired task space velocity $\dot{\mathbf{x}}_{Mnp}\in\mathbb{R}^n$ and we set $\mathbf{J}_{Mnp} = \mathbf{I}^{n\times n}$ to equally use all joints for this task. \textit{Mnp} defines an activation threshold $l_{Mnp} = 0.08$ from which $\eta_{Mnp}$ increases linearly to 1.0 at $l_{Mnp} = 0.04$.\newline\newline%
Similar to \textit{Motion}, the resulting control commands of \textit{Jla}, \textit{Coll} and \textit{Mnp} are scaled according to Eq.~(\ref{Eq:scaling}) to achieve zero-clamped velocities at the start and goal of a motion.%
\subsection{Metric-based Continuous Task Prioritization}\label{sec:Priorization}%
During motion, depending on the current robot and environment state, the priorities of tasks have to change for robust operation. For instance, when the robot is in the broader area of obstacles but not too close, \textit{Coll} is active but still of lower priority than reaching a defined goal. The closer the robot gets to obstacles, the higher the priority of \textit{Coll} over \textit{Motion}. Similarly, when not too close to the joint limits, \textit{Jla} might only be active in the nullspace of \textit{Motion}, but at some point, \textit{Jla} becomes of higher priority. \cite{Liu2015} formalizes this idea with the task prioritization matrix $\mathbf{A}$ that is also used in \cite{Dehio2019} and in our case $\mathbf{A}\in\mathbb{R}^{4\times 4}$, as we define four tasks. The elements in $\mathbf{A}$ take values in the range [0.0, 1.0] and define the priorities among the tasks, e.g. $a_{ij}$ defines the priority of task $j$ over task $i$. \cite{Liu2015} and \cite{Dehio2019} change these priorities with time, i.e. $a_{ij}(t)$, whereas we change the priorities depending on the importance metrics, i.e. $a_{ij}(\eta)$:%
\begin{equation}%
    \mathbf{A} = \begin{pmatrix}%
                0 & 1-\eta_{Coll} & 1-\eta_{Coll} & 1-\eta_{Coll}\\%
                \eta_{Coll} & 0 & 1-\eta_{Jla} & 1-\eta_{Jla}\\%
                \eta_{Coll} & \eta_{Jla} & 0 & 1-\eta_{Motion}\\%
                \eta_{Coll} & \eta_{Jla} & \eta_{Motion} & 0%
    \end{pmatrix}\label{Eq:A}%
\end{equation}%
The diagonal entries are zero. Otherwise, a task would be projected into its own nullspace, resulting in a deactivated task. Further, $a_{ij} = 1 - a_{ji}$ gives a proper priorization among two tasks.\newline%
We choose the following order of the rows/columns of~$\mathbf{A}$: \textit{Coll} in first row; \textit{Jla} in second row; \textit{Motion} in third row; \textit{Mnp} in fourth row. This order of the rows already defines one aspect of the prioritization: Let us assume that all~$\eta$ are 1.0, i.e. all tasks are fully active. In this case, the task in the first row (here \textit{Coll}) will use the full available space because its command will not be projected in any of the nullspace of the tasks of lower priority. This information is stored in~$a_{12},a_{13},a_{14}$. Similarly, in this case, that all tasks are fully active, the second row of~$\mathbf{A}$ indicates that the task of the second row (here \textit{Jla}) will be fully projected in the nullspace of the first task (here \textit{Coll}), which is indicated by~$a_{21}$. But it will not be projected into the nullspace of its lower tasks (here \textit{Motion} and \textit{Mnp}), as indicated by~$a_{23},a_{24}$. This approach avoids meaningless prioritization, which is outlined in \cite{Dehio2019} as an open topic: In contrast to the approach in \cite{Liu2015,Dehio2019}, Eq.~(\ref{Eq:A}) avoids meaningless prioritizations like $Coll > Motion > Jla > Coll$.% 
%
%
% ===============
% --- Results ---
% ===============
%
\section{Results}\label{sec:Results}%
\begin{figure}%
	\centering%
	\fontsize{8pt}{11pt}\selectfont%
	\subfloat[Dual-Arm Test Scenario\label{fig:Exp}]{%
		\includegraphics[width=0.45\columnwidth, height=3cm]{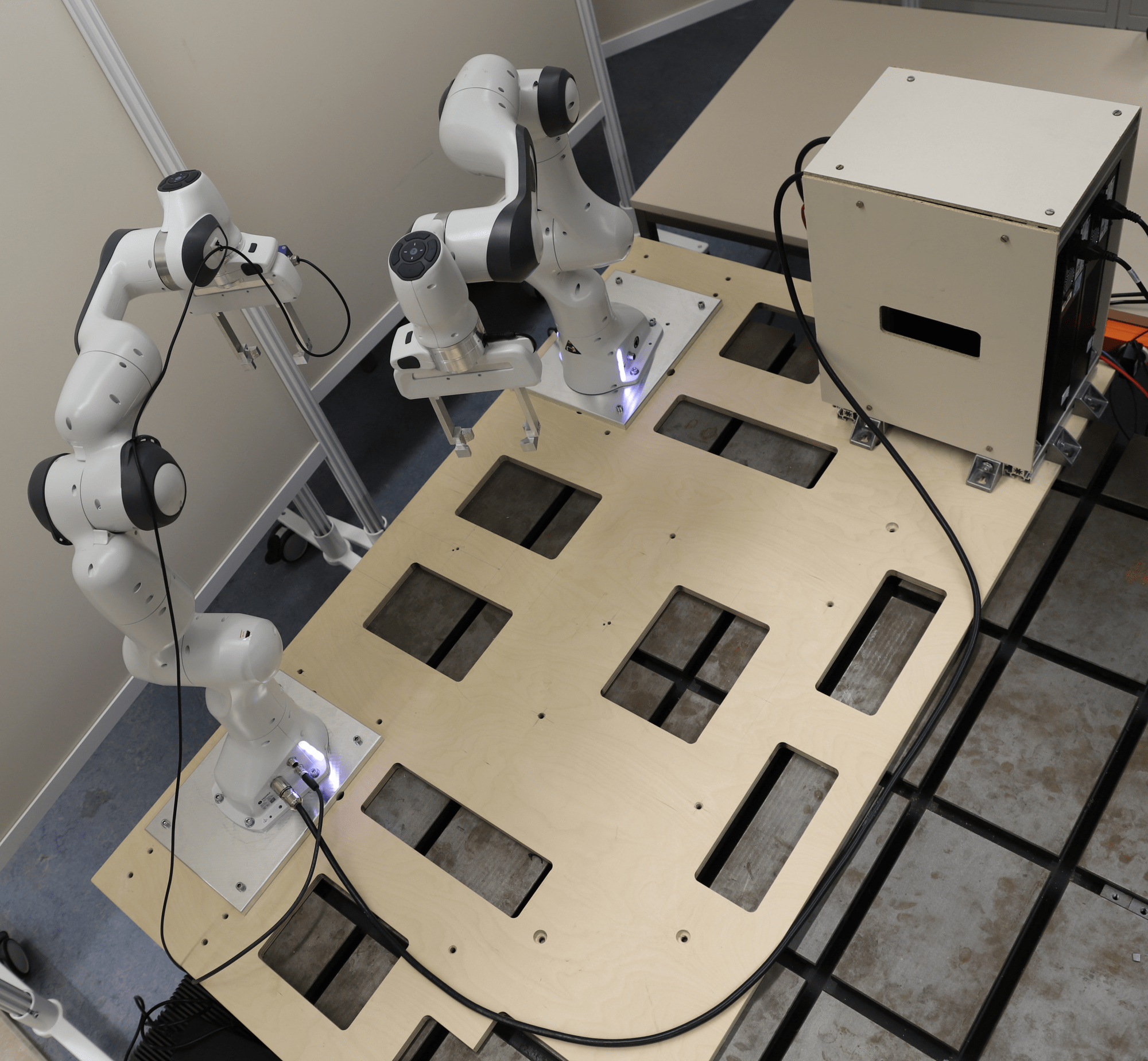}%
	}\hfil%
	\subfloat[\textit{SSV} Collision Model\label{fig:Ssv}]{%
	    \includegraphics[width=0.45\columnwidth, height=3cm]{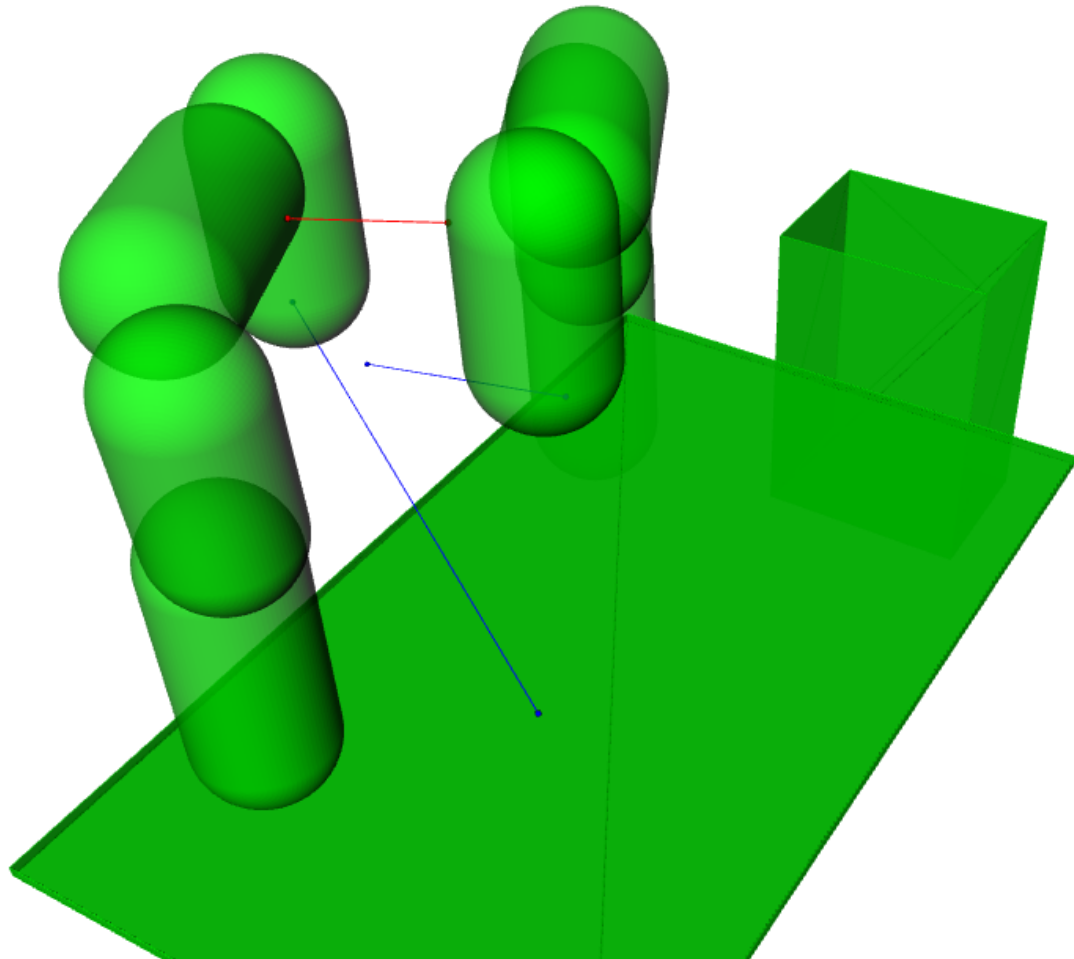}%
	}%
	\caption{Experimental validation scenario. In Figure~\ref{fig:Ssv}, the first robot is on the left and the second robot is on the right. The blue lines show the potential field-based PTP motion of the two end effectors if there were no other tasks. The red line shows the closest distance between a robot and another robot/the environment}%
	\label{fig:Setup}%
\end{figure}%
We implemented our proposed approach in the C++ framework that controls our two \textit{Franka Emika} robots shown in Figure~\ref{fig:Setup}. The robots have seven DoF each. We use the joint velocity control interface that expects continuous control commands. In the evaluation scenario, both robots move from an initial configuration to a Cartesian goal pose. The accompanying video shows the experiment and is available at: \href{https://youtu.be/uWFl824h1Po}{https://youtu.be/uWFl824h1Po}. During the motion, the robots get close to each other such that \textit{Coll} is activated, and they get close to their joint limits such that \textit{Jla} is activated. \textit{Mnp} tries to keep an advantageous configuration in motion segments of low manipulability.\newline%
\begin{figure}%
	\centering%
	\includegraphics[width=\columnwidth]{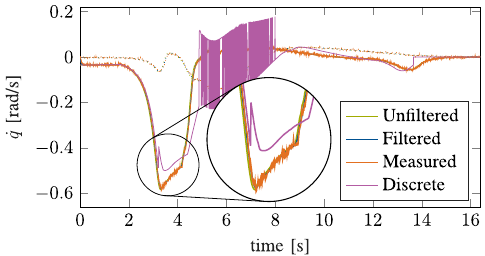}%
	%\resizebox{\columnwidth}{!}{%%
	%	\input{figures/tikz/JointVelocities.tex}%
	%}%
	\caption{Dotted and solid lines show the velocity control commands for joint~1 and~4, respectively. Green lines show the unfiltered commands, blue lines show the filtered commands, and orange lines show the measured velocities. The pink line shows the discontinuous control, expemplarily for joint~4, that would result from an approach with discrete task prioritization}%
	\label{fig:Qdot}%
\end{figure}%
\begin{figure*}%
	\centering%
	\fontsize{8pt}{11pt}\selectfont%
	\subfloat[Importance metrics $\eta$\label{fig:Eta}]{%
		\includegraphics[width=1.3\columnwidth]{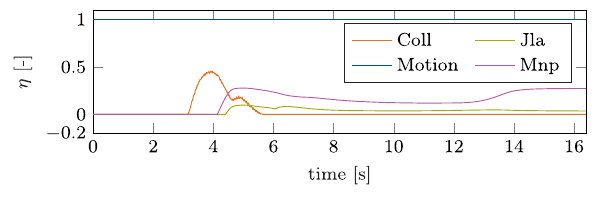}%
	}\hfil%
	\subfloat[$a_{ij}$ for $i\equiv\text{\textit{Motion}}$. Pink and blue lines overlap\label{fig:MotionPrio}]{%
		\includegraphics[width=1.3\columnwidth]{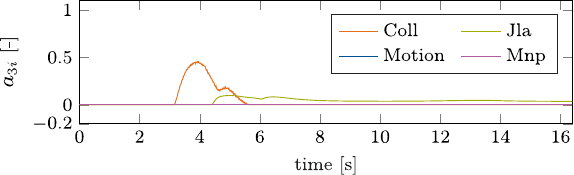}%
	}\hfil%
	\subfloat[Computed \textit{Motion} command according to Eq.~(\ref{Eq:FirstOrder}) and projected \textit{Motion} command after nullspace projection according to line 13 in Alg.~\ref{Alg:HierarchicalControl} for joints~1 (solid) and~4~(dashed)\label{fig:CompVsProj}]{%
		\includegraphics[width=1.3\columnwidth]{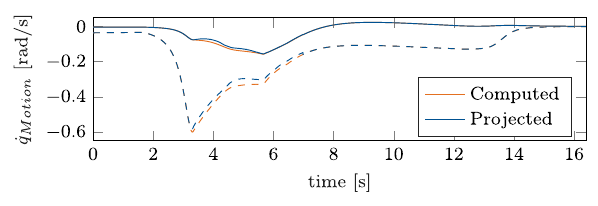}%
	}\hfil%
	\subfloat[Resulting command and projected task contributions for joint~1. The black line is the global command that corresponds to the dotted line in Figure~\ref{fig:Qdot}\label{fig:TaskContributions}]{%
		\includegraphics[width=1.3\columnwidth]{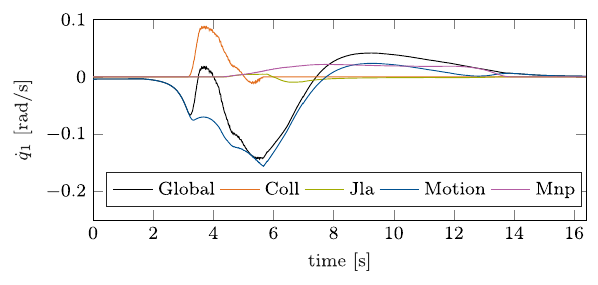}%
	}
	\caption{Task (de)activation throughout the motion of the experiment available at \href{https://youtu.be/uWFl824h1Po}{https://youtu.be/uWFl824h1Po}}%
	\label{fig:Priorization}%
\end{figure*}%
For the first robot, Figure~\ref{fig:Qdot} shows the resulting control command $\dot{\mathbf{q}}_d$, exemplarily for joint~1 (dotted) and~4 (solid), i.e. $\dot{\mathbf{q}}_d$ includes all projected tasks according to line~13 in Alg.~\ref{Alg:HierarchicalControl}. Our computed control in green is continuous with close-to-zero velocities at start and goal. The velocities are not exactly zero at start because we allow a small velocity such that the robot starts moving. The control interface of the robots expects continuous signals also on acceleration level. Therefore, the computed $\mathcal{C}^0$-continuous velocity commands are filtered with a PT1, however, Figure~\ref{fig:Qdot} shows that the filtering changes the values only slightly, i.e. the filtered values in blue are close to the unfiltered values in green. That is, our filter is not used to smooth a discontinuous velocity signal. Such a discontinuous signal would result from an approach that does not change the priorities based on continuously evolving importance metrics but that changes the priorities abruptly. The pink line in Figure~\ref{fig:Qdot} shows this for joint~4, with the same application scenario as for the experiments. Note that the abrupt change of priorities also leads to instabilities in task rearrangement, i.e. frequent changes. The control interface of our robots does not accept such a discontinuous signal, so in the accompanying video, we show this case only in simulation but for the same application as in the experiments.\newline%
For the first robot, Figure~\ref{fig:Priorization} shows the effect of task (de)activation throughout the robot motion in the experiment. Figure~\ref{fig:Eta} shows that while the robot approaches the other robot, \textit{Coll} gets more important because the distance decreases, i.e. $\eta_{Coll}$ increases. Also, for \textit{Jla} and \textit{Mnp}, the priorities increase as the robot gets closer to the joint limits and configurations of low manipulability. As explained earlier, $\eta_{Motion}=1.0$ throughout the motion. Depending on the evolution of $\eta_k$, the priorities in $\mathbf{A}$ change according to Eq.~(\ref{Eq:A}). Figure~\ref{fig:MotionPrio} shows the row in~$\mathbf{A}$ that corresponds to \textit{Motion}, i.e. it shows the priorities of other tasks over \textit{Motion}. As outlined, the diagonal terms of $\mathbf{A}$ are~0.0, thus, $a_{33}=0.0$ (blue line in Figure~\ref{fig:MotionPrio}). As further outlined, the order of the rows of $\mathbf{A}$ already defines one aspect of the prioritization: The priority of \textit{Coll} and \textit{Jla} over \textit{Motion} change according to their $\eta_k$. In contrast, \textit{Mnp} is in a higher-indexed row in $\mathbf{A}$ than \textit{Motion}, and thus its priority over \textit{Motion} is defined by $\eta_{Motion}$ and not by $\eta_{Mnp}$ (see Section~\ref{sec:Priorization}). $\eta_{Motion}$ is set to~1.0 throughout the whole motion and thus the pink line is also at~0.0.\newline%
For joints~1 (solid) and~4 (dashed) of the first robot, Figure~\ref{fig:CompVsProj} shows that, after other higher-prioritized tasks are activated, only a part of the desired computed \textit{Motion} command (orange line) gets propagated to the final joint control command (blue line).\newline%
For joint~1 of the first robot, Figure~\ref{fig:TaskContributions} shows how much each task contributes to the final joint velocity control command. When the robot is still far from obstacles, \textit{Motion} is the only active task: For this period, the final control command (black line) overlaps with the \textit{Motion} command (blue line). As soon as \textit{Coll} gets activated, \textit{Motion} becomes less important, and the final control command starts to deviate. Similar observations can be made for \textit{Jla} and \textit{Mnp}.\newline%
The experiment is a contribution on its own: The control of several tasks is computationally expensive. \cite{Liu2015} and~\cite{Dehio2019} show \textit{GHC} and \textit{DynGHC} only in simulation. However, both outline in their conclusions that future work should include \textit{"the reduction of the computational cost of GHC to achieve real-time control of complex robots"}~\cite{Liu2015} and that future work \textit{"is necessary to evaluate the approach on real robots for useful multi-task applications"}~\cite{Dehio2019}. We achieve experiments on hardware though our four defined tasks include two computationally expensive ones: \textit{Coll} requires distance computations that often make up 90\,\% of the computation in motion planning, and \textit{Mnp} includes, as outlined, the elaborate computation of the gradient according to~\cite{Baur2012}. Further, Alg.~\ref{Alg:HierarchicalControl} itself is computationally expensive in real-time control, e.g. it includes the high-dimensional augmented Jacobian $\mathbf{J}_{aug}$ that stacks all task Jacobians. To achieve an efficient computation of \textit{Coll}, we model the robot and the environment with \textit{swept sphere volumes (SSV)}, i.e. simplified geometries that provide fast distance computation (see Figure~\ref{fig:Ssv}). For this, we use the open-source \textit{broccoli} C++ library~\cite{Seiwald2022}. With this implementation, we can compute Alg.~\ref{Alg:HierarchicalControl} at 1\,kHz, which is the real-time control of the robots' control interface.
%
%
% ===================
% --- Conclusions ---
% ===================
\section{Conclusions}\label{sec:Conclusions}%
This work contributes to hierarchical control. First, Section~\ref{sec:Theory} reviews weighted pseudo-inverses of Jacobians. The discussion itself is not novel, however, its comprehensiveness is. We clearly distinguish control methods (velocity, acceleration, and torque control), task mapping, and nullspace projection. The conclusion of the first part is: The standard approach in Eq.~(\ref{Eq:ASC}) only minimizes the tracking error of the secondary task. It can be enhanced with a scalar parameter to also consider velocity-squared norms. However, in the latter case, both weighting matrices must be equally shaped. Our approach allows to consider kinetic energy and task tracking error, i.e. it supports differently shaped weighting matrices. Second, Section~\ref{sec:HierarchicalControl} presents hierarchical velocity control. The algorithm for velocity-based control and applying importance metrics for task rearrangement are novel. We demonstrate the real-time capability of our formulation in experiments, which is also not achieved in \cite{Liu2015} and \cite{Dehio2019}.\newline%
Our approach still suffers from local minima, which is a well-known problem in potential field-based approaches. Proposed solutions are the definition of potential fields with no local minima, e.g. the \textit{Brushfire Algorithm} \cite{Choset2005}, or random walks to get out of local minima. Further, given several task commands, our algorithm computes $\mathcal{C}^0$-continuous velocity commands. However, this is only guaranteed if the individual tasks also compute $\mathcal{C}^0$-continuous commands. For instance, if the closest collision pair in \textit{Coll} abruptly changes, the task computes a discontinuous desired velocity that also propagates to the final command. Filtering techniques may overcome this issue. Finally, in Section~\ref{sec:MappingAndProjection}, we derive an approach for a hierarchical control that allows to trade off energy consumption and tracking error. We implement this approach into the generic formulation of Alg.~\ref{Alg:HierarchicalControl}, and we prove the correct formulation by comparing the simulation result of Alg.~\ref{Alg:HierarchicalControl} with a simulation based on an analytical closed-form solution. However, we do so only for two priority levels. In applications with more than two active tasks, we do not have a reference to verify the implementation.\newline%
In this work, we proposed our hierarchical control approach, benchmarked its characteristics against other methods in 100 simulation scenarios, and validated its real-time capability in an experiment.%
%
%
%%%%%%%%%%%%%%%%%%%%%%%%%%%%%%%%%%%%%%%%%%%%%%
%%%%%%%%%%%%%%%%% Backmatter %%%%%%%%%%%%%%%%%
%%%%%%%%%%%%%%%%%%%%%%%%%%%%%%%%%%%%%%%%%%%%%%
%
\backmatter
%
%\mathbfhead{Supplementary information}
%\mathbfhead{Acknowledgments}
%
%
%%%%%%%%%%%%%%%%%%%%%%%%%%%%%%%%%%%%%%%%%%%%%%
%%%%%%% Statements and Declarations %%%%%%%%%%
%%%%%%%%%%%%%%%%%%%%%%%%%%%%%%%%%%%%%%%%%%%%%%
%
\section*{Statements and Declarations}
% The following statements must be included in your submitted manuscript under the heading 'Statements and Declarations'. This should be placed after the References section. Please note that submissions that do not include required statements will be returned as incomplete.
%
\textbf{Funding - }The authors declare that no funds, grants, or other support were received during the preparation of this manuscript.\newline%
\textbf{Competing Interests - }The authors have no relevant financial or non-financial interests to disclose.\newline%
% Author Contributions. For instance:
% Conceptualization: [full name], …; Methodology: [full name], …; Formal analysis and investigation: [full name], …; Writing - original draft preparation: [full name, …]; Writing - review and editing: [full name], …; Funding acquisition: [full name], …; Resources: [full name], …; Supervision: [full name],….
\textbf{Author Contributions - }\textbf{Jonas Wittmann}: Conceptualization, Methodology, Software, Validation, Formal Analysis, Investigation, Data Curation, Writing - Original Draft, Writing - Review and Editing, Visualization, Supervision. \textbf{Daniel Hornung}: Conceptualization, Methodology, Software, Validation, Investigation, Data Curation, Writing - Review and Editing. \textbf{Korbinian Griesbauer}: Methodology, Writing - Review and Editing. \textbf{Daniel Rixen:} Conceptualization, Resources, Writing - Review \& Editing, Supervision, Project Administration, Funding Acquisition.\newline%
\textbf{Ethics approval - }Not applicable.\newline%
\textbf{Consent to participate - }Not applicable.\newline%
\textbf{Consent to publish - }Not applicable.\newline%
%
%
%%%%%%%%%%%%%%%%%%%%%%%%%%%%%%%%%%%%%%%%%%%%%%
%%%%%%%%%%%%%%%%%% Bibliography %%%%%%%%%%%%%%
%%%%%%%%%%%%%%%%%%%%%%%%%%%%%%%%%%%%%%%%%%%%%%
%
\bibliography{sn-bibliography}% common bib file
\end{document}